\documentclass[11pt]{article}
\usepackage{coling2018}
\usepackage{times}
\usepackage{url}
\usepackage{amsmath}
\usepackage{amsfonts}
\usepackage{booktabs}
\usepackage{cases}
\usepackage{color}
\usepackage{graphicx}
\usepackage{latexsym}
\usepackage{multicol}
\usepackage{subcaption}
\usepackage{xspace}

\usepackage{xcolor}

\newcommand{\myquote}[1]{``#1''}

\newcommand{\aps}{'s\xspace}
\newcommand{\ap}{'\xspace}

\newcommand{\eqspace}{\vspace{-0.5cm}}

\newcommand{\ie}{{\it i.e.}\xspace}

\newcommand{\mysec}{{Section}\xspace}

\newcommand{\deen}{English/German\xspace}
\newcommand{\enge}{German/English\xspace}
\newcommand{\en}{English\xspace}

\newcommand{\de}{German\xspace}
\newcommand{\wmtfour}{WMT\ap2014\xspace}
\newcommand{\wmtfive}{WMT\ap2015\xspace}
\newcommand{\wmtsix}{WMT\ap2016\xspace}

\newcommand{\attentivelm}{\emph{Attentive} RNN-LM\xspace}
\newcommand{\attentivelms}{\emph{Attentive} RNN-LMs\xspace}
\newcommand{\attentivenmt}{\emph{Attentive} NMT\xspace}

\newcommand{\indomain}{\emph{in-do\-main\nocorr}\xspace}

\title{Exploring the Use of Attention within an Neural Machine Translation Decoder States to Translate Idioms}

\author{
  Giancarlo D. Salton \and Robert J. Ross \and John D. Kelleher\\
  Applied Intelligence Research Centre \\
  School of Computing \\
  Dublin Institute of Technology \\
  Ireland \\
  {\tt \{giancarlo.salton,robert.ross,john.d.kelleher\}@dit.ie}
 \\}
\date{}

\begin{document}
\maketitle
\begin{abstract}
  Idioms pose problems to almost all Machine Translation systems. This type of language is very frequent in day-to-day language use and cannot be simply ignored. The recent interest in memory augmented models in the field of Language Modelling has aided the systems to achieve good results by bridging long-distance dependencies. In this paper we explore the use of such techniques into a Neural Machine Translation system to help in translation of idiomatic language.
\end{abstract}

\section{Introduction}
\label{sec:intro}

The recent interest in Deep Learning (DL) research has resulted in ideas originating in the DNNs field being applied to Machine Translation (MT), resulting in the development of a new area of research called Neural Machine Translation (NMT). The basic idea of NMT is to apply two different Recurrent Neural Networks (RNNs), which are a specific type of DNNs for processing sequences, in the so-called Encoder/Decoder framework \cite{sutskever:2014,cho:2014}. The first RNN, called the \textit{encoder}, is trained to compress the input sentence, written in the source language, into a distributed representation (\ie, a fixed-size vector of real numbers). The second RNN, called the \textit{decoder}, is trained to take that distributed representation and decompress it (word-by-word) into the output sentence, written in the target language.

A number of extensions to the basic encoder-decoder architecture have been proposed. For example, both \newcite{bahdanau:2015} and \newcite{luong:2015} integrate attention mechanisms into the NMT architecture to improve the alignment of information flow between the encoder and the decoder at each step in the generation of the translation. However, a potentially interesting (and as yet under-explored) extension to the encoder-decoder framework is the use of attention within the decoder itself. The decoder is essentially an RNN language model (RNN-LM) that is conditioned on a representation of the input sentence generated by an encoder RNN. Consequently, the quality of a translation is dependent on the power of the RNN-LM that is used by the NMT architecture. Within language model research there have been a number of augmentations proposed to improve the performance of RNN-LM, such as the \newcite{daniluk:2017}, \newcite{merity:2017} and \newcite{salton:2017b}. However, to-date these innovations in the design of RNN-LMs have not been integrated into NMT decoders. 

Many of the augmentations to RNN-LMs are focused on improving the ability of the language models to span long-distance dependencies. Consequently, an interesting case-study to use to explore the potential of using attention within an NMT Decoder is the translation of idioms. In general, the performance of MT systems degrade when processing idioms \cite{vilar:2006,salton:2014a}. Furthermore, the inclusion of an idiomatic phrase within a translation introduces a break into the evolving context of the RNN-LM decoder that the decoder must bridge once the idiom has been generated. Hence, improving the ability of the RNN-LM decoder to span long-distance dependencies may result in improved performance on the translation of idioms.

In this paper we propose an extension to NMT that uses a RNN-LM which has been augmented with memory and an attention mechanisms as a decoder. We evaluate this model on a number of different datasets, a present and analysis of it performance on sentences that include idioms. We begin by introducing NMT in more detail and outlining previous work in the field in \mysec \ref{sec:related}. In \mysec \ref{sec:model} we describe the modifications we made to a baseline NMT architecture in order to integrate an \attentivelm into the architecture as its decoder (we refer to this augmented NMT architecture as an \attentivenmt). We then describe a set of experiments that evaluated the performance of an \attentivenmt over benchmark datasets, first when translating idioms and then when translating regular language \mysec \ref{sec:experiments}. In \mysec \ref{sec:analysis} we present an analysis of the performance of an \attentivenmt system compared to a set of baselines systems. Finally, in \mysec \ref{sec:conclusions} we draw our conclusions.

\section{Neural Machine Translation}
\label{sec:related}
\begin{figure*}[!b]
\centering
\includegraphics[width=0.7\linewidth]{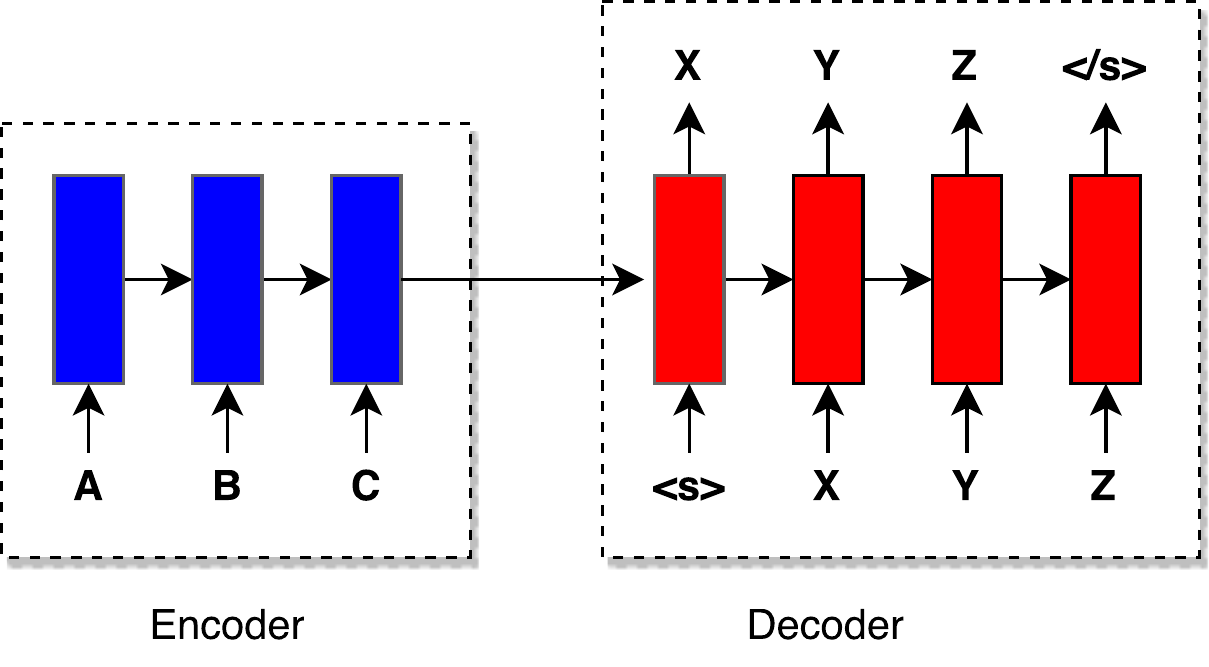}
\caption{The Encoder-Decoder architecture. The rectangles inside the box on the left represent the Encoder RNN unfolded over time and the rectangles inside the box on the right represent the Decoder RNN, also unfolded over time. The output of the last step of the Encoder RNN is the distributed representation of the source sentence. The first input for the Decoder RNN is the start-of-sentence ( \textless s\textgreater) symbol of the output sentence and the Decoder RNN\aps first hidden state is set to be the distributed representation of the input sentence. At each timestep $t_j$ the Decoder RNN emits a symbol (word) that will serve as input to the Decoder RNN at timestep $t_{j+1}$. The Decoder RNN performs computations until the \textless /s\textgreater\space symbol is emitted. \label{fig:seq2seq}}
\end{figure*}

Figure \ref{fig:seq2seq} informally outlines the basic NMT model. Formally, let $S = \{w_{1}^{s}, \dots, w_{L_S}^{s}\}$ and $T = \{w_{1}^{t}, \dots, w_{L_T}^{t}\}$ represent represent the source sentence $S$  and the target sentence $T$ respectively. $L_S$ and $L_T$ denote the lengths of $S$ and $T$ respectively (note that $L_S$ might be different from $L_T$). Also let $n_e$ denote the number of hidden units in the \textit{Encoder} and $n_d$ denote the number of hidden units in the \textit{Decoder}.

\noindent \textbf{Encoder}. The \textit{Encoder} compresses information about $S$ into a distributed representation $c$ using a RNN. The computation of $c$ involves iterating over the following equation:

\vspace{-0.2cm}
\begin{align}
\mathbf{h}_{i}^{e} &= f(w_{i}^{s}, \mathbf{h}_{i-1}^{e})
\end{align}\label{eq:rnn_encoder}

\noindent where $f$ is a non-linear function; and $\mathbf{h}_{i}^{e} \in \mathbb{R}^{n_e}$ (also called the \textit{Encoder hidden state}) is the output of $f$ at each iteration $i$. The \textit{Encoder} then outputs its last \textit{hidden state} to be the representation $\mathbf{c}^e$\footnote{Note that $\mathbf{c}^e \in \mathbb{R}^{n_e}$)}:

\vspace{-0.2cm}
\begin{align}
\mathbf{c}^e &= \mathbf{h}_{L_s}^{e}
\end{align}\label{eq:distributed_representation}

\noindent .

\noindent \textbf{Decoder}. The \textit{Decoder} is often trained to predict the next word $w_{j}^{t}$ given $\mathbf{c}^e$ and all previous $w_{<j}^{t}$ (\ie, all words of previous timesteps to $w_{j}^{t}$). Therefore, the \textit{Decoder} is understood to define a probability over the translation $T$ using the (ordered) conditionals:

\vspace{-0.2cm}
\begin{align}
P(T) &= \prod_{j=1}^{L_T}P(w_{j}^{t} | \{ w_{1}^{t}, \dots, w_{j-1}^{t} \}, \mathbf{c}^e)
\end{align}\label{eq:decoder_probabilities}

\noindent where the probability $P(w_{j}^{t} | \{ w_{1}^{t}, \dots, w_{j-1}^{t} \}, \mathbf{c}^e)$ is defined as:

\vspace{-0.2cm}
\begin{align}
P(w_{j}^{t} | \{ w_{1}^{t}, \dots, w_{j-1}^{t} \}, \mathbf{c}^e) &= g(w_{j-1}^ {t}, \mathbf{h}_{j}^{d})
\end{align}\label{eq:rnn_decoder}

\noindent where $g$ is a non-linear function; and $\mathbf{h}_{j}^{d} \in \mathbb{R}^{n_d}$ (also called the \textit{Decoder hidden state}) is the output of $g$ at each iteration $j$.

The \textit{Sequence-to-Sequence Learning} (\textit{Seq2seq}) model proposed by \cite{sutskever:2014} closely follows the \textit{Encoder-Decoder} approach using Long-Short Term Memory (LSTM) \cite{hochreiter:1997} units. A single \textit{Seq2seq} model has achieved results close to the state-of-the-art SMT systems \cite{sutskever:2014}. Despite being relatively simple in comparison with other NMT systems, the ensemble of \textit{Seq2seq} models using LSTM units still achieves the state-of-the-art results for French to English translation \cite{sutskever:2014}. Nevertheless, a negative point of these models is the fact they are difficult to train given the required number of hidden layers for each neural network and the size of the models. In fact, these models can easily have more then 26 billion parameters.

An alternative NMT model incorporating an attention mechanism was proposed by \newcite{bahdanau:2015}. Their model, called  \textit{NMT with Soft Attention} (or \textit{NMT with Global Attention}), also builds upon the \textit{Encoder-Decoder} approach and adds a small Neural Network that learns which part of the encoded distributed representation of the source sentence to pay attention to at the different stages of the decoding process. This model is more complex then Seq2seq but requires a smaller number of hidden layers and parameters. This reduction in hidden layers and parameters is due to the use of Gated Recurrent Units (GRU) \cite{cho:2014} in the hidden layers of this model. This approach was the first pure NMT system to win a Machine Translation Shared Task in the Workshop on Statistical Machine Translation (WMT) \cite{bojar:2015}.

More recently, \newcite{luong:2015} proposed the \textit{NMT with Local Attention} model, building upon the ideas of both \textit{Seq2seq} and \textit{NMT with Global Attention}. In this work, \newcite{luong:2015} also use two stacks of LSTM units (similar to \textit{Seq2seq}) and includes two feedforward networks. The first network is trained to predict a fixed-size window over the distributed representation of the source sentence. The second network is trained to learn which part of the predicted window to pay attention to at different stages of the decoding step, similar in spirit to \textit{NMT with Global Attention}. The \textit{NMT with Local Attention} model is currently the state-of-the-art for translating from English into German \cite{luong:2015}. In addition, \newcite{luong:2015} also experimented with the use of stacks of LSTM units together with a \textit{Global attention} mechanism (more general than the original), with slightly worse results than the \textit{Local Attention} model.

\section{\attentivenmt}
\label{sec:model}
According to \newcite{sutskever:2014}, the \textit{decoder} of a NMT system is essentially a RNN-LM conditioned on an input sequence encoded by another RNN (\ie, the \textit{encoder}). To date, the extensions to the \textit{Seq2seq} architecture have focused on improving the representation generated by the encoder (for example, using bi-directional inputs \cite{bahdanau:2015}) or using attention (either local or global) to provide an alignment mechanisms between the decoder and the encoder. In this paper we adopt a different approach and explore how extensions to the decoder can help with the translation of idioms. Specifically, we experiment with using at language model that integrates an attention mechanism. In recent years a number of augmented language models have been proposed \cite{daniluk:2017,grave:2017,merity:2017,salton:2017b}. For the purposes of this work, any of these augmented LM systems could be used as a decoder in the NMT system. However, for our experiments we chose to use the attentive Language Model architecture of \newcite{salton:2017b}. Our motivation for using this model was the relative  compactness of the architecture in terms of parameter size, its suitability for sentence based processing, and ease of implementation. In order to adapt the \attentivelm of \newcite{salton:2017b} to work as the decoder of a NMT system, we use the NMT architecture described in \newcite{luong:2015}.

\newcite{luong:2015} adds a memory buffer where all the hidden states $h_{1}^{e}, \dots, h_{L_S}^{e}$ of the \textit{encoder} RNN are stored. The authors then modify Eq. \ref{eq:distributed_representation} so that a new representation $\mathbf{c}^{e}$ for the input sequence is generated at each timestep $t$ using an attention based model. More formally, generating $c_{t}^{e}$ involves iterating over the following equations:

\eqspace
\begin{align}
\mathbf{c}_{t}^{e} = \sum_{i=1}^{L_S} a_i \mathbf{h}_{i}^{e}
\label{eq:enc_context}
\end{align}
\vspace{-0.4cm}
\begin{align}
a_i = \frac{exp(score(\mathbf{h}_{i}^{e}, \mathbf{h}_{t}^{d}))}{\sum_{j=1}^{L_S} exp(score(\mathbf{h}_{j}, \mathbf{h}_{t}^{d}))}
\label{eq:enc_softmax}
\end{align}
\begin{numcases}{score(\mathbf{h}_{i}^{e}, \mathbf{h}_{t}^{d})=}
	 \mathbf{h}_{i}^{e}\odot\mathbf{h}_{t}^{d} & \text{dot}\label{eq:casedot}\\
	 \mathbf{h}_{i}^{e}\odot \mathbf{W}_{a} \mathbf{h}_{t}^{d} & \text{general}\label{eq:casegeneral}\\
	 \mathbf{W}_{a}[\mathbf{h}_{i}^{e};\mathbf{h}_{t}^{d}] & \text{concat}\label{eq:caseconcat}
\end{numcases}
\vspace{-0.3cm}

\noindent where $\odot$ is a dot product and $\mathbf{W}_a$ is a matrix of parameters.

The vector $c_{t}^{e}$ is then merged with the current state $\mathbf{h}_{t}^{d}$ by means of a concatenation layer

\eqspace
\begin{align}
\mathbf{h}_{t}^{\prime\prime} = tanh(\mathbf{W}_c[\mathbf{h}_{t}^{d};c_{t}^{e}] + \mathbf{b}_t)
\label{eq:concat_nmt}
\end{align}
\eqspace

\noindent where $\mathbf{h}_{t}^{\prime\prime}$, in the original model, is then passed to the softmax layer to make the prediction for the next word.

As we are interested in the simple models, we choose to integrate the \attentivelm with \textit{single score} of \newcite{salton:2017b}. This model performs reasonably well despite being simple. This model generates a context vector ($\mathbf{c}_{t}^{d}$) based on previous hidden states of the RNN by calculated a weighted sum of these hidden states (see Equation \ref{eq:context}). The weights used in the calculation of the summation represents the attention ($a_i$) the model pays to each of the past states ($\mathbf{h}_{i}^{d}$). This context vector is then used to bring forward, to the prediction step, past information processed by the model. More formally, the context  $\mathbf{c}_{t}^{d}$ for the \textit{decoder} is generated by iterating over the following equations:

\eqspace
\begin{align}
\mathbf{c}_{t}^{d} = \sum_{i=1}^{t-1} a_i \mathbf{h}_{i}^{d}
\label{eq:context}
\end{align}
\vspace{-0.4cm}
\begin{align}
a_i = \frac{exp(score(\mathbf{h}_{i}^{d}, \mathbf{h}_{t}^{d}))}{\sum_{j=1}^{t-1} exp(score(\mathbf{h}_{j}^{d}, \mathbf{h}_{t^{d}}))}
\label{eq:softmax}
\end{align}
\vspace{-0.4cm}
\begin{align}
score(\mathbf{h}_{i}^{d}) &= \mathbf{v}_{s} \odot tanh(\mathbf{W}_s \mathbf{h}_{i}^{d})\label{eq:single}
\end{align}
\eqspace

According to \newcite{salton:2017b}, that score reflects the ``relevance'' of the state $\mathbf{h}_{i}^{d}$ to the current prediction. The context $\mathbf{c}_{t}^{d}$ is merged with the current state $\mathbf{h}_{t}^{d}$ using a concatenation layer:

\eqspace
\begin{align}
\mathbf{h}_{t}^{\prime} = tanh(\mathbf{W}_c[\mathbf{h}_t;\mathbf{c}_t] + \mathbf{b}_t)
\label{eq:concat}
\end{align}
\eqspace

The next word probability is computed using using Eq.\ref{eq:probability}:

\eqspace
\begin{align}
p(w_t|w_{<t}, x) = softmax(\mathbf{W}_{soft} \mathbf{h}_{t}^{\prime}  + \mathbf{b})
\label{eq:probability}
\end{align}
\eqspace


In order to integrate the \attentivelm as a decoder within \newcite{luong:2015} NMT architecture,  we add a concatenation layer so that we may merge $\mathbf{h}_{t}^{\prime\prime}$ with $\mathbf{h}_{t}^{\prime}$ to bring froward information from previously predicted words:

\eqspace
\begin{align}
\mathbf{h}_{t}^{\dag} = tanh(\mathbf{W}[\mathbf{h}_{t}^{\prime\prime};\mathbf{h}_{t}^{\prime}] + \mathbf{b})
\label{eq:concat_attentive_nmt}
\end{align}
\eqspace

\noindent $\mathbf{h}_{t}^{\dag}$ is then passed to the softmax layer (Eq. \ref{eq:probability}) to make the next prediction. In all equations above, $\mathbf{W}_{*}$ indicate a matrix of parameters, $\mathbf{v}_{*}$ indicate a vector of parameters and $\mathbf{b}_{*}$ indicates a bias vector.

\section{NMT Experiments}
\label{sec:experiments}
We are interested on the effectiveness of the \attentivenmt when translating idioms. We describe the experiment on the translation of idioms from \enge and from \deen in \mysec \ref{sub:idioms}. To get some intuition on how the model performs on regular language, we also tested the model over benchmark datasets as described in \mysec \ref{sub:wmt}. We present and discuss the results obtained by our models in \mysec \ref{sub:results}

\subsection{Idioms Experiment}
\label{sub:idioms}

\textbf{Datasets}. As our idiom dataset for the \deen language pair (in both directions), we used the dataset of \newcite{fritzinger:2010}. This dataset was extracted from EUROPARL and from a \de newspaper corpus\footnote{\textit{Frankfurter Allgemeine Zeitung}}. The dataset consists of sentences containing one of 77 \de preposition+noun+verb triples. Each sentence is annotated with syntactic and morphological features and has one of four labels: \textit{idiomatic}; \textit{literal}; \textit{ambiguous}; or \textit{error} (which indicates parsing or extraction errors). Around $95\%$ of the dataset consists of idioms and the remaining $5\%$ consists of the other 3 labels. Therefore, we considered only the 3,050 sentences extracted from EUROPARL given that it is a bilingual corpus. From these sentences, we randomly selected 2,200 sentences to build a test set and used the remaining 850 sentences as part of our training data so as to ensure that there were idioms in the training data.

Given that the EUROPARL is part of the training dataset used in the \wmtsix, we used that dataset (after the pre-processing performed on the EUROPARL part of that data as described above) to train ours and the baseline models.

\textbf{\attentivenmt}. We use \attentivenmt models of similar sizes in comparison to those proposed by \newcite{luong:2015}. More specifically, the encoder is an RNN composed of four layers of 1,000 LSTM units and the decoder is the \attentivelms (also composed of four layers of 1,000 LSTM units). We also applied attention over the encoder outputs using the \myquote{dot} and \myquote{general} content functions (Eq. \ref{eq:casedot} and Eq. \ref{eq:casegeneral} respectively) as described in \newcite{luong:2015}. In total, we have trained four models for each direction (eight in total).

We optimize the model using ADAM \cite{kingma:2014} with a learning rate of 0.0001 and mini-batches of size 128 to minimize the average negative log probability of the target words. We train the models until we do not get any improvements on the negative log probability over the development set with an early stop counter of 5 epochs\footnote{In general, the average number of epochs is around 14, including the 5 epochs of patience.}. Once the model runs out of patience, we rollback its parameters and use the model that achieved the best performance on the validation set to obtain the translations. We initialize the weight matrices of the network uniformly in [$-0.05$, $0.05$] while all biases are initialized to a constant value of $0.0$. We also apply $50\%$ dropout to the non-recurrent connections and clip the norm of the gradients, normalized by mini-batch size, at 5.0. In all our models (similar to \cite{press-wolf:2016}), we tie the matrix $\mathbf{W}$ in Eq. \ref{eq:probability} to be the embedding matrix (which also has 1,000 dimensions) used to represent the input words. We remove all sentences in the training set larger than 50 words and we pad all sentences shorter than 50 words with a special symbol so they will all have the same size. We use a vocabulary of the 50,000 most frequent words in the training set including three symbols to account for the padding of shorter sentences, the end of sequence and OOV words respectively.

\textbf{NMT Baseline.} We used the model of \newcite{sennrich-wmt:2016}\footnote{Code available at \url{http://data.statmt.org/rsennrich/wmt16_systems/}} as our NMT baseline. In fact, this model is equal to the model of \newcite{bahdanau:2015}  apart from the fact that it uses BPE to reduce the size of the vocabulary. In this model we used word embeddings of 500 dimensions and 1,024 GRU units in both encoder and decoder. The recurrent weight matrices were initialized  as random orthogonal matrices; the bias vectors were initialized at 0; the parameters of the attention layer (the encoder attention layer) were sampled from a Gaussian distribution with $0$ mean and variance $0.001^{2}$; and the remaining weight matrices were sampled from a Gaussian distribution with $0$ mean and variance $0.01^{2}$; the model was trained using Adadelta \cite{zeiler:2012} with a learning rate of $0.0001$ (and a \myquote{patience} of 10 epochs) and the norm of the gradients (normalized by mini-batch size) were clipped at $1.0$.

\textbf{SMT Baseline.} We trained a PBSMT system using the Moses toolkit \cite{koehn:2007} with its baseline settings. We used a 5-gram LM trained with the KenLM \cite{heafield:2011} toolkit with \textit{mo\-di\-fied-kne\-ser-ney} smoothing. In this case, we used the \textit{newstest2013} (3K sentences) set to tune this model.

\subsection{Regular Language Experiments}
\label{sub:wmt}

To get some intuition on their performance over regular language, we evaluate our models over two benchmark datasets, the \wmtfour and \wmtfive test sets. We used the same \attentivenmt and baseline models trained for the idioms experiment described in \mysec \ref{sub:idioms}.

\begin{table*}[]
 \caption[Results in terms of BLEU and TER score on the idioms test set for the \deen language pair in both directions]{Results in terms of BLEU and TER score on the idioms test set for the \deen language pair in both directions. \myquote{Dot} and \myquote{General} refer to the content function applied to compute the attention over the encoder states. Larger BLEU and smaller TER scores indicates better performance.}
 \setlength\tabcolsep{24pt}
 \begin{tabular*}{\textwidth}{lcc}
   \toprule
     Model & \multicolumn{2}{c}{Idioms dataset} \\
     \cmidrule{2-3}
     & BLEU & TER \\
   \midrule
     English/German (EN/DE) \\
   \midrule
   NMT                                             & 14.4           & 72.4          \\
   SMT                                             & 15.9           & 68.7          \\
   Dot \& \attentivenmt      & 19.5           & 66.9          \\
   General \& \attentivenmt  & \textbf{19.9}  & \textbf{64.1} \\
   \midrule
     German/English (DE/EN) \\
   \midrule
   NMT                                             & 20.1           & 64.2 \\
   SMT                                             & \textbf{24.0}  & \textbf{60.6} \\
   Dot \& \attentivenmt      & 13.9           & 74.1 \\
   General \& \attentivenmt  & 14.3           & 73.8 \\
 \end{tabular*}\label{table:results_nmtende_idioms}
\end{table*}

\begin{table*}[]
  \caption[Results in terms of BLEU and TER score on the \wmtfour and \wmtfive test sets]{Results in terms of BLEU and TER score on the \wmtfour and \wmtfive test sets. \myquote{Dot} and \myquote{General} refer to the content function applied to compute the attention over the encoder states. Larger BLEU and smaller TER scores indicates better performance.}
  \setlength\tabcolsep{13pt}
  \begin{tabular*}{\textwidth}{lcccc}
    \toprule
      Model & \multicolumn{2}{c}{\wmtfour} & \multicolumn{2}{c}{\wmtfive} \\
      \cmidrule{2-3} \cmidrule{4-5}
      & BLEU & TER & BLEU & TER \\
    \midrule
      English/German (EN/DE) \\
    \midrule
    SMT                                             & 14.8  & 69.4  & 17.3  & \textbf{66.1}  \\
    NMT                                             & 13.6  & 72.3  & 13.8  & 72.4  \\
    Dot \& \attentivenmt      & 15.3  & \textbf{67.1}  & 17.2  & 64.8  \\
    General \& \attentivenmt  & \textbf{16.1}  & 68.7  & \textbf{17.8}  & 66.2 \\
    \midrule
      German/English (DE/EN) \\
    \midrule
    SMT                                             & \textbf{21.6}  & \textbf{58.6}  & \textbf{22.9}  & \textbf{56.9} \\
    NMT                                             & 17.9  &	63.3  &	17.7	& 63.9 \\
    Dot \& \attentivenmt      & 14.2  &	70.1  & 14.8  & 68.1 \\
    General \& \attentivenmt  & 13.9  & 70.8  & 14.6  & 68.4 \\
  \end{tabular*}\label{table:results_nmtende}
\end{table*}

\subsection{Results}
\label{sub:results}

Table \ref{table:results_nmtende_idioms} presents the results in terms of BLEU and TER scores over the dataset of idioms of \newcite{fritzinger:2010}. Given that this idiom corpus was extracted from the data used to train all models, we can consider it an \indomain test set. In the \deen direction, both of our models performed better than all the baselines, including the SMT, in terms of both BLEU and TER. Our models had a similar performance when the BLEU scores are considered, but there was a difference of almost 3 TER points between our best and our worst models. However, in the \enge direction, both of our models had the worst performance among all models in terms of both BLEU and TER scores. The SMT system performed best in this language direction, achieving almost 20 BLEU points more than our best model.

Table \ref{table:results_nmtende} show the results of our experiments in the \deen language pair in terms of BLEU and TER scores. Although the baseline NMT is the model that won the \wmtsix shared task on the \deen language pair, on both directions, the results obtained in our experiments were below the SMT baseline for both datasets (\wmtfour and \wmtfive) and both metrics (BLEU and TER). However, the entries submitted by \newcite{sennrich-wmt:2016} were the results obtained by an ensemble of these baseline models, including reranking and other post-processing steps on the output. As we are interested in the comparisons of single models, we do not apply the same steps as they did.

In the \en/\de direction, the General \& \attentivenmt achieved the highest BLEU score on both \wmtfour and \wmtfive test sets. It is noteworthy that the best model in terms of BLEU is not the best model in terms of TER. Using this metric to score the systems, the \attentivenmt and \emph{dot} encoder content function had a better performance on \wmtfour while the baseline SMT performed better on the \wmtfive. None of the \attentivenmt models had the same level of performance in the \enge direction as the baselines, \ie NMT and SMT.

All results presented in this section were tested for significance using the method of \newcite{koehn:2004} and we found all ($p \ll 0.05$).

%
%

\section{Analysis of the Models}
\label{sec:analysis}
The results obtained by our \attentivenmt are mixed. The \attentivenmt models improved the translations from \en into \de, including the translation of idioms. Given the fact that the \de language has long-distance dependencies in its syntax, this result may be an indication that our model helps in bridging those dependencies by bringing forward past information that may have faded from context at the target side with the \textit{smoothing effect} as shown in the language modeling experiments of \newcite{salton:2017b}. However, that improvement is not observed when \en is the target side. A mitigating point for consideration here is that, as shown by \newcite{jean-wmt:2015}, it is recognized that NMT models often struggle when translating into English (in comparison to translating into other languages) and even the inclusion of an attention module to score past states of the decoder may not overcome this challenge for NMT.

However, this still leaves open the question of why the baseline NMT system outperforms the \attentivenmt models when translating into English. Our hypothesis for what causes this is that the attention mechanism within the decoder interferes with the encoder-decoder alignment mechanism. When translating languages, there are interactions between input and output words that are highly informative to the system when making the next prediction and that are captured via word alignments \cite{koehn:2010}. By introducing a module to score states of the decoder and subsequently merge this information to the alignments calculated by the attention module over the encoder states (as we do in Eq. \ref{eq:concat_attentive_nmt} after the computation of the alignments by the with the \textit{Encoder} states) we are introducing a bias towards the decoder states and, thus, we may be weakening the information carried about input/output alignments. If the model is not robust enough to balance this trade off, it will fail and produce poor translations, a fact that is observed in several of our models.

\section{Conclusions}
\label{sec:conclusions}
In this paper we have studied the inclusion of the attention module over the decoder states of a NMT system. Although this type of attention module has achieved good performance in language modeling, such improvement were limited in the NMT setting. We have shown that, although our systems do not achieve state-of-the-art results for \deen direction, they are on par with single models that compose the ensembles that won the \wmtsix shared task and with baseline SMT systems. Despite the fact that the results are mixed, we have demonstrated that in some cases the representations built by the attention module also improves the translation of idioms, especially when the target language has long-distance dependencies such as the case for \de.

In future work we plan to investigate the usage of other architectures of memory-agumented RNN-LMs with the decoder of the NMT system. We also plan to integrate the computation of the attention over the encoder states and the attention over the decoder states into a single step.


\section*{Acknowledgements}

The acknowledgements should go immediately before the references.  Do
not number the acknowledgements section. Do not include this section
when submitting your paper for review.

\bibliographystyle{acl}
\bibliography{coling2018}

\end{document}